\documentclass[letterpaper, 10 pt, conference]{ieeeconf}  
\usepackage{blindtext, graphicx}
\usepackage{tabularx}
\usepackage[table,xcdraw]{xcolor}
\usepackage{adjustbox}
\usepackage{multirow}
\usepackage{amsmath}
\usepackage{caption}
\usepackage{authblk}
\usepackage{multicol}
\usepackage[tight,footnotesize]{subfigure}
\usepackage{cite}  

\IEEEoverridecommandlockouts
\title{\Large \bf A Next-Best-Smell Approach for Remote Gas Detection with a Mobile Robot}

\author{Riccardo Polvara$^{1}$$^{*}$, Marco Trabattoni$^{2}$$^{*}$, Tomasz Piotr Kucner$^{3}$, Erik Schaffernicht$^{3}$ \\ Francesco Amigoni$^{2}$ and Achim J. Lilienthal$^{3}$

\thanks{$^{1}$Dipartimento di Elettronica, Informatica e Bioingeneria, Politecnico di Milano,  20133, Milano, Italy }

\thanks{$^{2}$Autonomous Marine Systems Research Group, School of Marine Science and Engineering, Plymouth University, PL4 8AA, Plymouth, England.}

\thanks{$^{3}$Mobile Robotics \& Olfaction Lab, Center of Applied Autonomous Sensor Systems (AASS), \"Orebro
Universitet, SE-70182, \"Orebro, Sweden.}

\thanks{$^{*}$Both first and second author contributed equally and should be considered co-first authors.  {\tt riccardo.polvara@plymouth.ac.uk}  {\tt marco.trabattoni@mail.polimi.it}}
}

\begin{document}

\maketitle

\begin{abstract}
The problem of gas detection is relevant to many real-world applications, such as leak detection in industrial settings and landfill monitoring. Using mobile robots for gas detection has several advantages and can reduce danger for humans. In our work, we address the problem of planning a path for a mobile robotic platform equipped with a remote gas sensor, which minimizes the time to detect all gas sources in a given environment. We cast this problem as a coverage planning problem by defining a basic sensing operation -- a scan with the remote gas sensor -- as the field of ``view'' of the sensor. Given the computing effort required by previously proposed offline approaches, in this paper we suggest a online coverage algorithm, called \emph{Next-Best-Smell}, adapted from the Next-Best-View class of exploration algorithms. Our algorithm evaluates candidate locations with a global utility function, which combines utility values for travel distance, information gain, and sensing time, using Multi-Criteria Decision Making. In our experiments, conducted both in simulation and with a real robot, we found the performance of the Next-Best-Smell approach to be comparable with that of the state-of-the-art offline algorithm, at much lower computational cost.\\
\end{abstract}


%
\IEEEpeerreviewmaketitle

\section{Introduction}
\label{introduction}


\indent In recent years, mobile robot olfaction, a branch of robotics that combines gas sensors with the flexibility of mobile robots, attracted growing attention \cite{mobile}, since it allows robots to perform tasks (e.g., exploration, surveillance, search and rescue) in potentially hazardous conditions, such as in presence of a possible gas leak. The problem of detecting a gas source is of utmost interest especially in the case of methane leaks, since methane is an extremely flammable greenhouse gas and being able to detect a gas source is the first step towards sealing it. This interest has further increased with the advent of remote gas sensors which, in contrast to \textit{in situ} gas sensors, do not need to get in direct contact with gas and can sense gas remotely up to $30$ meters~\cite{gas1,gas2}. Examples are the Tunable Diode Laser Absorption Spectroscopy (TDLAS) sensors~\cite{tdlas1,tdlas2}.

\indent Compared to a sensor network, the use of a mobile robot equipped with a remote gas sensor represents a flexible solution, provided that the robot is able to autonomously plan and follow paths in the environment. Without assuming any \textit{a priori} knowledge on gas distribution, the problem of gas detection in a known environment can be reduced to that of efficiently covering the environment with the measurements of the remote gas sensor. An offline solution to this problem has been presented in~\cite{asif} that plans a complete tour in the environment so that all points are covered by the gas sensor.

\indent Our contribution, the \emph{Next-Best-Smell} approach, adapts an online Next-Best-View (NBV) method proposed for exploration of initially unknown environments that uses Multi-Criteria Decision Making (MCDM) \cite{mcdm} to choose the next location a robot should reach to perform a measurement with its remote gas sensor. The aim is to minimize the total time required to cover an environment and thus to find all gas sources in a target area. At each step, candidate locations are identified along the boundary between scanned and unscanned space. With MCDM, the robot evaluates the candidate locations according to an utility function that combines different criteria, for example the distance of the candidate location from the robot or the expected amount of new information acquirable from there.\\

\indent The results obtained in our simulation and real-world experiments show that the Next-Best-Smell algorithm achieves a fast coverage of 80-90\% of the environment, followed by a slow convergence to 100\%. We discuss how the parameters of the MCDM can be tuned to achieve faster convergence for different types of environments (e.g., narrow corridors, open spaces). The comparison with a state-of-the-art offline algorithm shows a good performance of the proposed online Next-Best-Smell approach.

The main strength of the online greedy algorithm that we propose is that it is computationally light, and thus scales favourably to large environments. The proposed approach is fast and, in contrast to offline approaches, can be used in unknown environments concurrently to a obstacle mapping method.
In short, our main original contribution is a greedy online coverage algorithm that performs comparably with an offline variant, but with greatly reduced computational costs.

The remainder of the article is organised as follow: in Section \ref{problem-def} and in Section \ref{nbs} we define the problem and introduce our algorithm, respectively. In Section \ref{experiments} we report the results of experiments, performed both in simulation and with a physical platform, and we compare our Next-Best-Smell approach against the work of Arain \textit{et al.} \cite{asif}. In Section \ref{state} we survey the related literature while in Section \ref{conclusion} we summarize our approach, highlighting the main concepts, its advantages, and future works.

\section{Problem Definition}\label{problem-def}




We assume that the task of gas detection is to be performed by a single robot. For the initial discussion, the environment is assumed to be known in advance and represented by an occupancy grid of identical squared cells, labeled as either free or obstacle cells.

The pose $p$ of the robot is defined as $p = (c,\theta)$.
The position is represented by the free cell $c$ in which the robot is currently located (we assume that the robot is always in the center of the cell). The orientation $\theta$ belongs to a finite set \textit{$\Theta$} of possible orientations, with values equally spaced in \textit{[0,$2\pi$)}.

\begin{figure}
\centering
\includegraphics[width=\linewidth]{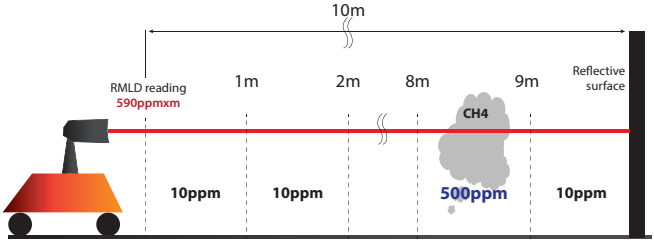}
\caption{The Remote Methane Leak Detector is a TDLAS sensor which reports the integral concentration of methane along its laser beam (parts per million x meter)}
\label{tdlas}
\end{figure}

The robot is assumed to move on the grid, that can be 4- or 8-connected, from a pose $p$ to a pose $p'$ without traversing obstacle cells. The robot can perform a \emph{sensing operation} to analyze the presence of gas in its current field of view. Gas sensing is carried out using an onboard TDLAS remote gas sensor that reports the integral concentration measurements of the target gas along a beam, even in presence of multiple airborne substances (see Fig. \ref{tdlas}). In order to scan a portion of the environment a pan-tilt unit is used to aim the sensor at a sequence of orientations, performing a sweep and thus scanning a circular sector of range \textit{r} and scan angle $\phi$. The scan angle $\phi$ defines the sweep that the gas sensor has to perform. The values of range and scan angle are physically restricted by the limit value $r_{max}$ and by the maximum opening angle $\phi_{max}$, respectively, due to sensor's physical constraints. The larger the scan angle $\phi$ the longer the time the sensing operation takes. Specifically, the time required for the sensing operation increases with the scan angle $\phi$ in a non-linear way due to the set and reset time required to correctly position the pan-tilt unit before and after the sweep, and is also influenced by the range used. On the GasBot platform \cite{bennetts}, performing a scan with a range $r$ = 5 meters and a sweep of $\phi$ = 45 degrees takes about 21 seconds, while performing a sweep of $\phi$ = 90 degrees takes about 36 seconds. A sensing operation is thus defined by a robot pose $p=(c,\theta)$, the radius \textit{r}, and the scan angle \textit{$\phi$}, that together define the \textit{Field of Smell} (FoS) of the robot, as the set of cells that are perceived from $p$.

\begin{figure} [tb]
    \centering
{\includegraphics[width=0.8\columnwidth, height = 5cm]{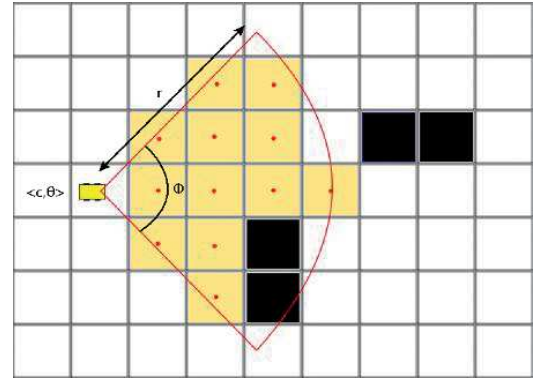}} \quad
\caption{Grid representing the environment. The highlighted cells are covered by the sensing operation performed at the robot's location.}
\label{grid}
\end{figure}

We assume that a free cell $c'$ is \textit{smellable} from a robot pose $p=(c,\theta)$ if the line segment spanning from the center of $c$ to the center of $c'$ does not intersect any obstacle cell and if the center of $c'$ is inside the circular sector centered in $p$ and defined by $r$ and $\phi$, where $\phi$ spans symmetrically around the robot heading $\theta$. This corresponds to the assumption that all obstacles fully occupy grid cells and that they are high enough to obstruct the line of sight of the remote gas sensor. We assume that any sensing operation performed from pose $p$ detects the presence of gas in all the cells smellable from $p$. We also say that the smellable cells have been scanned or covered (and we mark them accordingly on the grid map). Fig. \ref{grid} shows the representation of the environment and the overage of a basic sensing operation.

\indent The problem of planning a path for gas detection in a given environment is that of finding the optimal sequence of sensing operations $\langle ((c_{1},\theta_{1}),r,\phi_{1}),  ((c_{2},\theta_{2}),r,\phi_{2}), \ldots, ((c_{n},\theta_{n}),r,\phi_{n})\rangle$ to be performed in order to scan (cover) all the free cells of the environment (namely, each celle should be smelled from at least a $((c_{i},\theta_{i}),r,\phi_{i})$). Pose ($c_{1}$,$\theta_{1}$) is the starting pose of the robot in the environment and is not required to be equal to pose ($c_{n}$,$\theta_{n}$), namely we are looking for a path and not for a tour. Performance metrics for optimality include the number $n$ of sensing operations and the time to fully cover the environment, both to be minimized.

\section{The Next-Best-Smell Approach}\label{nbs}

\begin{figure}
\centering
\includegraphics[width = \linewidth]{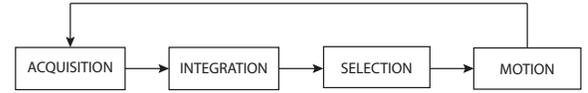}
\caption{Next-Best-View system: acquire with the TDLAS sensor a partial map of the surrounding environment, integrate it with the global map and update the list of candidate positions; adopt MCDM to select the best one and move to it in order to execute a new sensing operation. These steps are reiterated until full coverage.}
\label{nbv}
\end{figure}

In our approach to solve the above problem, the robot operates in an online fashion by iteratively perceiving the environment, updating the map with information from the most recent observation (presence of gas in scanned cells), deciding where to move next, and actually moving there (Fig. \ref{nbv}). The core of our approach is the decision step. When selecting the next pose to move to, we define \textit{candidate positions} the cells on the boundary between the portion of environment that has already been scanned and the one which has yet to be explored. For each of these candidate positions we consider multiple candidate robot poses, one for each orientation $\theta\in\Theta$.
In order to choose the best pose among the candidate ones, we define three \textit{criteria} for their evaluation:

\begin{itemize}
\item \textit{Travel distance}, computed as the distance between the current robot pose and the candidate pose using A* on the current map.
\item \textit{Information gain}, computed as the number of free unscanned cells that the robot will be able to perceive from the candidate pose.
\item \textit{Sensing time}, computed from the scan angle $\phi$ required to sense all the free unscanned cells visible from the candidate pose in the sensing operation. Note that this criterion is specific for olfaction and, in particular, for the TDLAS sensor we employ. The criterion has little relevance for sensors that operate in negligible time, like LIDARs for a obstacle mapping.
\end{itemize}

The values of information gain and sensing time are computed using \textit{ray casting}. For each free cell $a$ surrounding the candidate pose within $r$, a ray is cast from the candidate pose to the center of $a$. If the ray intersects an obstacle cell, it stops, while if it reaches the center of $a$, the cell is defined as smellable and the information gain count is increased. The angles of the first and last ray intercepting a smellable cell from a pose define $\phi$ for that pose, which is used to calculate the sensing time as the sum of the set up time and the sweep time, which we assume to depend linearly from $\phi$”. Note that our approach, given a pose, calculates the optimal value for $\phi$ in that pose, namely the smallest $\phi$ that smells all the unscanned cells at that pose. For each of these criteria, a utility value indicating how much a candidate pose satisfies the criterion is calculated; the value is normalized in order to obtain a number between 0 and 1, the higher the value the better the pose with respect to the criterion considered.

In order to select the best candidate pose, namely the one that best satisfies, in a balanced way, all the criteria, a global utility function combining these utility values is necessary. We define this function using the MCDM method, which has been proven useful for selecting candidate poses in exploration~\cite{mcdm}. This technique deals with problems in which a decision maker has to choose among a set of alternatives and its preferences depend on different, and sometimes conflicting, criteria. MCDM offers a principled way to combine criteria and to account for their dependencies. For example, two criteria might estimate similar features using two different methods. In this case, a relation of \textit{redundancy} can be modeled among them, and their overall contribution to the global utility should be less than their sum. On the other hand, two criteria might estimate two very different and complementary features, meaning that in general a candidate pose optimizing both of them is desirable. In this case, a relation of \textit{synergy} can be modeled among the criteria, and their overall contribution to the
global utility should be larger than their sum.

MCDM exploits an aggregation method called \textit{Choquet Fuzzy Integral} in order to account for the relations of redundancy and synergy when combining the utility values of criteria.
For our purposes (see~\cite{mcdm} for a complete description), we first need to introduce a function $\mu$ : $P(N) \to [0, 1]$, where $N$ is the set of criteria we consider ($|N|=3$ in our case) and $P(N)$ is the power set of $N$, with the following properties:

\begin{itemize}
\item $\mu(\{ \emptyset \}) = 0$,
\item $\mu$($N$) = 1,
\item if $A \subseteq B \subseteq N$, then $\mu(A) \le \mu(B)$. \end{itemize}

\noindent This means that $\mu$ is used to specify weights for each subset of criteria. The weights specified by $\mu$ capture the above relations among criteria: if two criteria $c_{1},c_{2}$ are redundant, then $\mu(\{c1,c2\}) < \mu(\{c1\}) + \mu(\{c2\})$, while if they are synergic $\mu(\{c1,c2\}) > \mu(\{c1\}) + \mu(\{c2\})$; in case $\mu(\{c1,c2\}) = \mu(\{c1\}) + \mu(\{c2\})$ we say that the criteria are \emph{independent}.
The global utility function of a candidate pose \textit{p} can then be computed as the discrete Choquet integral with respect to the fuzzy measure $\mu$ using the utilities of $p$ relative to the criteria:

\begin{center}\begin{equation}\label{choquet} \textit{$f(p)$ = $\displaystyle \sum_{j=1}^{|N|}(u_{(j)}(p) - u_{(j-1)}(p))\mu(A_{(j)})$}\end{equation} \end{center}
where \textit{$(j)$} indicates the j-th criterion in $N$ after permutation of criteria in order to have, for the candidate pose \textit{$p$}:\\
\begin{center} $u_{(1)}(p) \le ... \le u_{(|N|)}(p) \le 1$ \end{center}
(we assume $u_{(0)}(p)$ = 0) and the set $A$ is defined as:\\
\begin{center} $A_{(j)} = {i \in N | u_{(j)}(p) \le u_{(i)}(p) \le u_{(|N|)}(p)}$ \end{center}
%
%

\begin{table}[]
\resizebox{\linewidth}{!}{%
\begin{tabular}{|c|c|c|c|c|c|c|c|c|c|c|c|c|c|c|c|}
\hline
\rowcolor[HTML]{FFC702}
{\bf Configuration} & {\bf $x_{1}$} & {\bf$x_{2}$} & {\bf $x_{3}$} & {\bf $x_{1}$, $x_{2}$} & {\bf $x_{1}$,$x_{3}$} & {\bf $x_{2}$,$x_{3}$} & {\bf $x_{1}$,$x_{2}$,$x_{3}$}  \\ \hline
\cellcolor[HTML]{FFFFFF}{\bf A} & 1 & 0 & 0 & 1 & 1 & 0 & 1   \\ \hline
\cellcolor[HTML]{FFFFFF}{\bf B} & 0 & 1 & 0 & 1 & 0 & 1 & 1   \\ \hline
\cellcolor[HTML]{FFFFFF}{\bf C} & 0 & 0 & 1 & 0 & 1 & 1 & 1  \\ \hline
\cellcolor[HTML]{FFFFFF}{\bf D} & 0.333 & 0.333 & 0.333 & 0.766 & 0.766 & 0.766 & 1  \\ \hline
\cellcolor[HTML]{FFFFFF}{\bf E} & 0.6 & 0.2 & 0.2 & 0.9 & 0.9 & 0.5 & 1  \\ \hline
\cellcolor[HTML]{FFFFFF}{\bf F} & 0.428 & 0.428 & 0.144 & 0.956 & 0.672 & 0.672 & 1  \\ \hline
\cellcolor[HTML]{FFFFFF}{\bf G} & 0.2 & 0.6 & 0.2 & 0.9 & 0.5 & 0.9 & 1  \\ \hline
\cellcolor[HTML]{FFFFFF}{\bf H} & 0.144 & 0.428 & 0.428 & 0.672 & 0.672 & 0.956 & 1  \\ \hline
\cellcolor[HTML]{FFFFFF}{\bf I} & 0.2 & 0.2 & 0.6 & 0.5 & 0.9 & 0.9 & 1  \\ \hline
\cellcolor[HTML]{FFFFFF}{\bf J} & 0.428 & 0.144 & 0.428 & 0.672 & 0.956 & 0.672 & 1  \\ \hline
\cellcolor[HTML]{FFFFFF}{\bf K} & 0.5 & 0.5 & 0 & 1 & 0.6 & 0.6 & 1  \\ \hline
\cellcolor[HTML]{FFFFFF}{\bf L} & 0 & 0.5 & 0.5 & 0.6 & 0.6 & 1 & 1   \\ \hline
\cellcolor[HTML]{FFFFFF}{\bf M} & 0.5 & 0 & 0.5 & 0.6 & 1 & 0.6 & 1  \\ \hline
\end{tabular}}
\caption{Weights associated to each configuration: $x_{1}$ is the weight of \textit{information gain} criterion, $x_{2}$ is that of the \textit{travel distance}, and $x_{3}$ is that of the \textit{sensing time}.}
\label{criteria}
\end{table}

\begin{figure}
\centering
\includegraphics[width=0.7\linewidth]{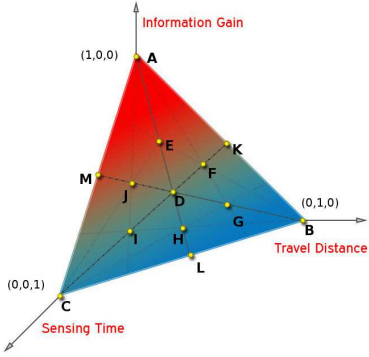}
\caption{On the simplex surface the points representing the configurations used are shown.}
\label{simplex}
\end{figure}

Finding the best weights for sets of criteria is a difficult but fundamental task that is often performed \emph{ad hoc} and empirically. However, when several criteria are involved, there are semi-automated methods, based on the \textit{Shapley value} concept inherited from game theory, that define constraints over some of the weights, by specifying bounds or exact values, and solve a corresponding linear program to find a feasible set of weights \cite{grabish3}. Given that we have only three criteria, we propose an approach that goes beyond the simple \emph{ad hoc} selection of weights. We set only on the weights of the three single criteria ($x_{1}$ is the weight of \textit{information gain} criterion, $x_{2}$ is that of the \textit{travel distance}, and $x_{3}$ is that of the \textit{sensing time}) and we model a slightly synergic relationship among them. We opted for synergy since the three criteria do not measure the same features and they are largely uncorrelated (for example, the distance of a candidate pose does not influence its expected information gain).
Our goal is to identify a set of weights that guarantee good results in terms of total coverage time.
We represent this optimization problem in a graphical way: due to the constraints
    \begin{equation} \label{eq:simplex}
     x_1 + x_2 + x_3 = 1
     \end{equation}
and considering $ x_1 \geq 0 $, $ x_2 \geq 0$, $ x_3 \geq 0 $ we can draw a simplex as the parameter space. We sample the parameter space as follows: the three vertices, the ortocenter, the middlepoint of each edge, and six symmetric points belonging to the three bisectors. All these selected points are reported in Table \ref{criteria} and shown in Fig. \ref{simplex}.
\indent Basically, they cover a number of potentially interesting situations for olfactory applications. For example, if we are interested in maximizing one criterion despite the others we can assign a value greater than 0.5 to that criterion and split the rest among the remaining, as Configurations E, G, and I.
Maximizing \emph{information gain}, \emph{travel distance}, and \emph{sensing time} criteria is expected to correspond to a robot behavior that quickly covers most of the environment, that travels short distances, and that completes the coverage with few sensing operations, respectively. Of course it is also possible to find a balance among multiple criteria: for example, Configuration F tries to optimize at the same time either the \textit{information gain} and the \textit{travel distance}, while Configuration D assigns equal importance to all the criteria. These configurations (sets of weights) are then evaluated experimentally on different maps.


\section{Experimental Results}\label{experiments}

We tested the proposed Next-Best-Smell approach in different scenarios, both in simulation and in the real world, using different maps and weight configurations.

\subsection{Parameters and Evaluation Metrics}

The gas detection process is affected by multiple parameters:
\begin{itemize}
\item The maximum range $r_{max}$ of the gas sensor (10 or 15 meters).
\item The scan angle $\phi$: due to the pan-tilt unit the gas sensor is mounted on, in our work we consider maximum opening angle $\phi_{max}$ = 180 degrees as the limit value for this parameter. $\phi$ is chosen for each sensing operation to only cover the unknown part of the map. This is a major difference with the offline approach in \cite{asif} that, to reduce the computing effort, uses a fixed value of $\phi$ for all the sensing operations.
\item The number of possible orientations $\Theta$ the robot can assume at a position: we assume either four (namely, N, E, S, W) or eight (plus N-E, N-W, S-E, S-W) orientations.
\item The amount of area to be covered (we consider $100\%$, namely full coverage).
\end{itemize}
The main metric considered in our experiments is the total coverage time, namely the sum of travelling and sensing time.
Coverage time is greatly affected by the number of times the robot stops in order to perform sensing operations, due to the time spent to correctly position the TDLAS for each sensing operation and the time required to perform a scan (up to dozens of seconds). However, the speed of a sensing operation could vary depending on the robotic platform used. For this reason we also report the total number of sensing operations.


\subsection{Simulations}

\begin{figure} [tb]
    \centering
{\includegraphics[width=.45\columnwidth, height = 4cm]{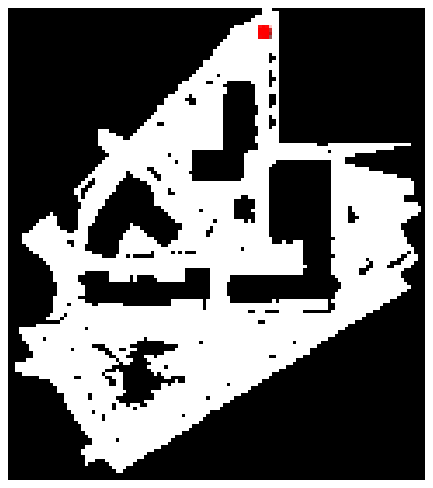}} \quad
{\includegraphics[width=.45\columnwidth, height = 4cm]{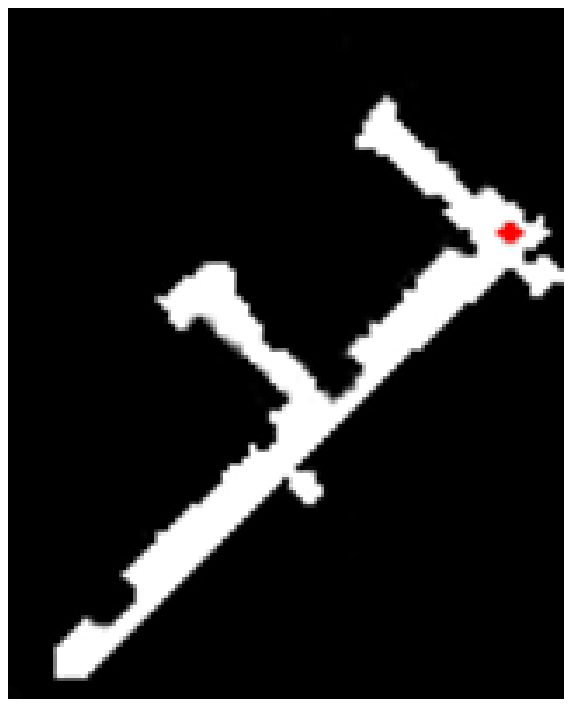}}
\caption{Maps used in the simulated experiments. On the left the Freiburg University map (approximately 6100 $m^{2}$), while on the right  the Teknikhuset corridor (approximately 560 $m^{2}$) of the \"Orebro Universitet. In both, the red dot represents the starting position of the robot.}
\label{fig:map}
\end{figure}

Multiple tests are run in simulation to evaluate our approach and also identify the best combinations of weights of criteria for best addressing the coverage problem, tuning different parameters.
The first environment we consider is that of the Freiburg University \cite{freiburg}, discretized in cells of one square meter, obtaining 6113 free cells. The map can be seen in Fig. \ref{fig:map}. A red dot identifies the starting position of the robot. The resolution adopted is adequate for gas detection, because gas forms larger plumes due to dispersal.

\begin{table}[]
\resizebox{\linewidth}{!}{%
\begin{tabular}{|c|c|c|c|c|c|c|c|c|}
\hline
\rowcolor[HTML]{FFCB2F}
Configuration & Coverage satisfied & Sensing ops &  Travel Time(s) &  Scanning time(s) & Total time(m) \\ \hline
A & yes & 110  & 2835.14 & 6018.28 & 147.55 \\ \hline
B & no & 207 & 5449.62  & 8200 & 227.49 \\ \hline
C & yes & 226 & 5896.08  & 6740.67 & 210.61 \\ \hline
D & yes & 163 & 3403.79  & 6045.3 & 157.48 \\ \hline
E & yes & 155 & 3380.36  & 5853.26 & 153.89 \\ \hline
F & yes & 157 & 2890.27  & 5899.98 & 146.50 \\ \hline
G & yes & 170 & 3225.13  & 6071.27 & 154.94 \\ \hline
H & yes & 193 & 4495.79  & 6505.33 & 183.35 \\ \hline
I & yes & 187 & 3826.46  & 6419.14 & 170.76 \\ \hline
J & yes & 171 & 3664.48  & 6005.91 & 161.17 \\ \hline
K & yes & 141 & 2233.14  & 6566.87 & 146.66 \\ \hline
L & no & 212 & 4797.31  & 7175.16 & 199.54 \\ \hline
M & yes & 171 & 3247 & 5809.77 & 150.94 \\ \hline
\end{tabular}}
\caption{Results for Freiburg University map for each considered configuration.}
\label{Freiburg}
\end{table}

\begin{figure}
    \centering
\includegraphics[width=\columnwidth, height=4cm]{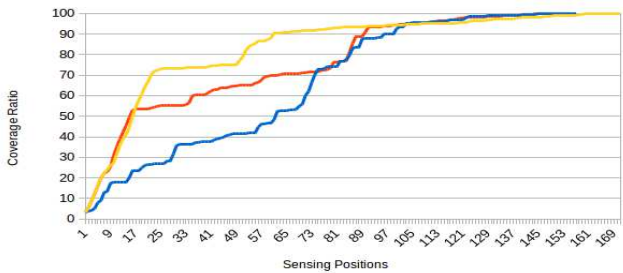}
\caption{Freiburg University's coverage: the graph represents the coverage ratio vs. sensing positions for configuration E(red), F(blue) and M(yellow).}
\label{freiburg-img}
\end{figure}

The results, presented in Table \ref{Freiburg}, show that best performance is obtained in situations in which \textit{information gain} and \textit{travel distance} are given more importance with large weights (Configurations F and K). Specifically, Configuration F produces the best result: 146.5 minutes as total coverage time, of which 48.2 minutes are used by the robot to move and the rest for scanning. Comparing this configuration with Configuration A, which also produces a good performance, it is evident that the robot stopped more often, however the sensing time is shorter. This is sound with the dynamic selection of scan angle $\phi$ and with weights assigned to criteria, since Configuration A considers only \textit{information gain}, while Configuration F puts emphasis on \textit{information gain} and \textit{travel time}, but also considers \textit{sensing time}.

Fig. \ref{freiburg-img} reports the coverage's rate with Configuration F: it is interesting to notice that the robot explored 4868 cells, corresponding to 80\% of the total number (6113), with 85 sensing positions, a bit more than half of the final number (157).
This behaviour is consequence of our greedy online approach which tends to maximize locally, initially discarding those poses with a small \textit{information gain} that, in most of the cases, correspond to corners.
Configurations B and L do not complete the coverage: 5 cells in the first case and 8 in the second are not scanned due to our ray casting implementation that considers as scanned only those cells whose center is inside the FoS.

\begin{figure}
    \centering
\includegraphics[width=\columnwidth, height=4cm]{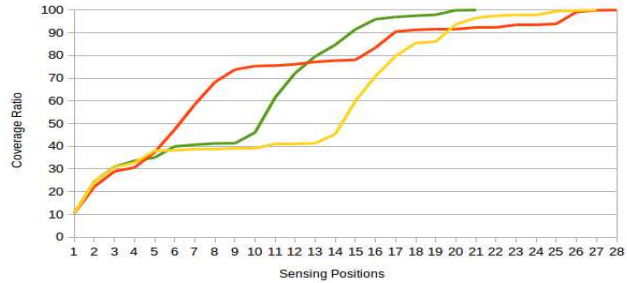}
\caption{Teknikhuset corridor's coverage: the represents the coverage ratio vs. sensing positions with configuration A(green), E(red) and M(yellow).}
\label{Teknihuset-img}
\end{figure}

Other tests are run on the Teknikhuset's corridor's map, shown in Fig. \ref{fig:map}. The results are reported in Table \ref{Teknihuset}.
We built the map with the physical robot. Then, we cleared the map from unreachable cells behind glass surfaces or under stairs and we discretized it in a grid with cells of half meter per side. Configuration M, shown in Fig. \ref{Teknihuset-img}, represents our best performance, scanning 730 cells over 916, corresponding to 80\%, with 17 sensing operations.

From the results obtained we can say that the choice of the weight configuration is related to the structure of the environment. In situations like the Freiburg example, with large open spaces, configurations that privilege \textit{travel distance} over \textit{sensing time} (F, G, K) are preferred because it is highly probable that scanning will be carried out with maximum opening angle and thus the \textit{sensing time} criterion is less meaningful. On the other hand, in environments with tight structured spaces as in the Teknikhuset corridor, it is more convenient to employ configurations that emphasize the \textit{sensing time} (I, J, M), because using only the angle strictly required for perceiving a new portion of map can improve significantly the performance by reducing time spent in sensing operations. In general, we found \textit{information gain} to be the most important criterion, since maximizing it leads to focus on covering new cells. For this reason, configurations of the upper part of the simplex reported in Fig. \ref{simplex} are often a good choice. In particular, a conservative solution such as Configuration E, which assigns a large weight to \textit{information gain} and smaller, equal weights to the other two criteria leads to good coverage times in both types of environment, as shown in Tables \ref{Freiburg} and \ref{Teknihuset}. For this reason, in situations in which the map of the environment is not provided (experiments have been done only in simulation) choosing configurations similar to E tends to result in good coverage time.


\begin{table}[]
\resizebox{\linewidth}{!}{%
\begin{tabular}{|c|c|c|c|c|c|c|c|c|}
\hline
\rowcolor[HTML]{FFCB2F}
Configuration & Coverage satisfied & Sensing ops & Travel Time(s)  & Scanning time(s) & Total time(m) \\ \hline
A & yes & 21 & 358.32  & 1107.79 & 24.44 \\ \hline
B & no & 43 & 528.15  & 1801.08 & 47.62 \\ \hline
C & yes & 44 & 522.64  & 1167.18 & 28.16 \\ \hline
D & yes & 40 & 730.29  & 1259.65 & 33.17 \\ \hline
E & yes & 28 & 541.18  & 1099.44 & 27.34 \\ \hline
F & yes & 39 & 509.84  & 1258.9 & 29.48 \\ \hline
G & yes & 39 & 1142.58  & 1261.35 & 40.07 \\ \hline
H & yes & 40 & 862.05  & 1205.72 & 34.46 \\ \hline
I & yes & 40 & 476.03  & 1106.57 & 26.38 \\ \hline
J & yes & 40 & 442.09  & 1197.23 & 27.32 \\ \hline
K & yes & 32 & 573.55  & 1334.52 & 31.80 \\ \hline
L & yes & 40 & 807.51  & 1185.90 & 33.22 \\ \hline
M & yes & 27 & 412.84  & 967.84 & 23.01 \\ \hline
\end{tabular}}
\caption{Results for Teknikhuset corridor map for each considered configuration.}
\label{Teknihuset}
\end{table}

\begin{figure}
\centering
\includegraphics[width=0.8\linewidth]{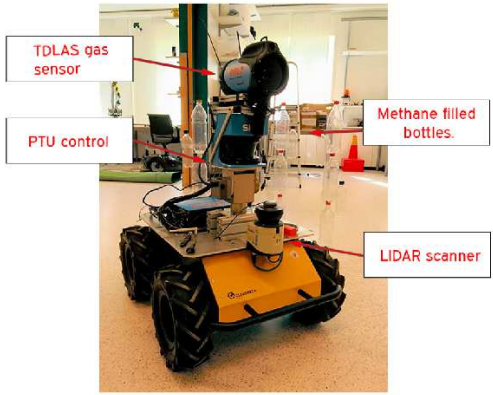}
\caption{Gasbot \cite{bennetts} equipped with a LIDAR scanner, a pan-tilt unit and a TDLAS sensor for remote gas detection. Methane leaks are simulated in the environment with four transparent bottles filled with the gas.}
\label{husky}
\end{figure}

\subsection{Real World Experiments}

\begin{figure}
\centering
\includegraphics[width=\linewidth]{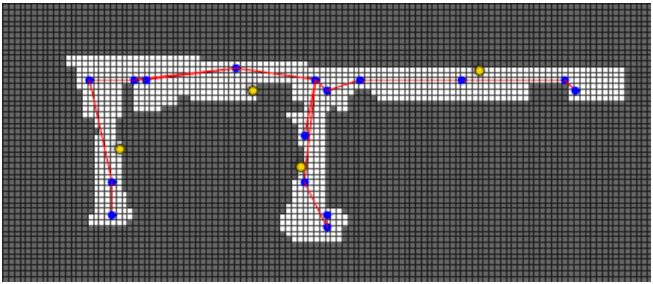}
\caption{Real-world experiment: the Teknikhuset corridor's map used in which blue dots represent the sensing positions of the robot and the yellow dots show the methane filled bottles.}
\label{real_sensing}
\end{figure}

We also performed experiments in the Teknikhuset's corridor with a physical platform, the GasBot \cite{bennetts}, a Husky A200 robot by Clearpath Robotics. As shown in Fig. \ref{husky} it is equipped with a LIDAR sensor and a Remote Methane Leak Detector (RMLD) on a pan-tilt unit. The RMLD uses a tunable infrared diode for open path optical absorption measurements and is selective to methane. Since releasing methane in public buildings is dangerous, we used transparent bottles filled with methane as sources detectable by the RMLD. Hence, the gas distribution is static and not affected by dispersal.
The experiment was run with the same map used in the simulation after removing all cells that are unreachable due to kinematics and physical constraints of the platform. In this way, we reduce the number of free cells from 916 to 896. The coverage was completed in 27.72 minutes and sensing positions are reported in Fig. \ref{real_sensing}.
During this experiment the robot used out-of-the-box navigation and localization available in ROS. These results are in line with what we expected from simulation: the Next-Best-Smell algorithm in simulation with the same map and the same configuration resulted in an estimated coverage time of 23.79 minutes. The  additional 4 minutes are due to errors introduced by the actual localization and navigation procedures. Additional sensing or navigation actions were executed by the robot to compensate.

Our gas maps just represent which cells have been perceived and which have not. However, \textit{a posteriori} analysis of the log files and direct observation during experiments suggest that the detected peaks of gas concentration are close to the positions of the bottles (a quantitative assessment is beyond the scope of this paper).

\subsection{Comparison with Offline Approach}

In order to evaluate the trade-off between efficiency of coverage and computing time of our online approach, we compare our results with those obtained with the offline approach of~\cite{asif}.
Comparison is carried out in simulation in two different experiments. In the first one, the Teknikhuset corridor's map is used, while, for the second one, we generate a set of random grids, with a size ranging from 3x3 to 90x90 cells in order to replicate the experiments of \cite{asif}. Ten grids are generated for each size. In each grid, 10\% of the cells are randomly marked as obstacles.
The differences between our online approach and the offline one are measured by the number of sensing operations (that, as we have seen, directly influences total coverage time) and time required to compute the covering path (i.e., sequence of sensing operations).

\begin{figure}
    \centering
\includegraphics[width=\columnwidth, height=4cm]{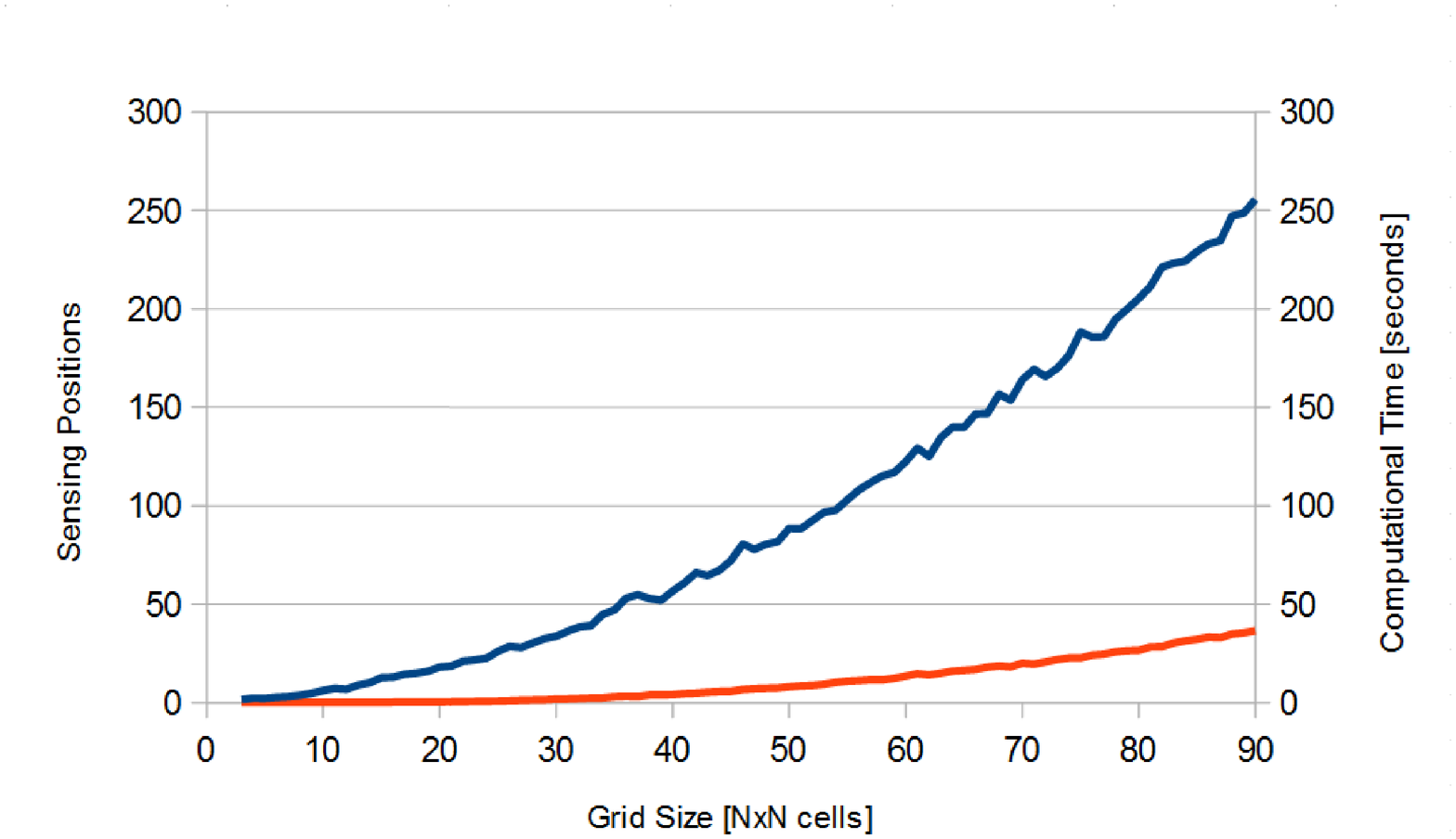}
\caption{Results obtained in simulation with random grids: the graph shows the average number of sensing operations (blue line) and computing time required to produce the solution (orange line).}
\label{graph_grids}
\end{figure}

Results regarding the first metric on the Teknikhuset's map are reported in Table \ref{comparison}.
In \cite{asif}, a configuration with $\phi_{max}=90$ degrees and range $r = 10$ meters for the gas sensor is used, with 4 possible orientations for each sensing operation, obtaining a total of 17 sensing operations and a total coverage time of 18.35 minutes. With the same parameters (Next-Best-Smell 90/4 in the table), our algorithm completes in 40 sensing positions, with a total coverage time of 27.78 minutes.
The difference in the number of sensing operations is expected, since the offline approach, exploiting the knowledge of the whole environment, is able to find a set of sensing poses (and a corresponding coverage path) closer to the optimal one. On the other hand, its greedy nature prevents our NBS approach from finding the minimum set of sensing poses.
This is further confirmed by tests on randomly generated grids (see Fig.~\ref{graph_grids}), using 4 possible orientations, $r = 30$ meters, and $\phi_{max}=180$ degrees, where the number of sensing operations obtained with our approach is roughly double that obtained in \cite{asif} experiments with the same parameters, reaching a total average of about 250 sensing positions in the 90x90 grid case.

However, the computational lightness of our algorithm permits us to use more fine grained orientations in our experiments, as shown in Table \ref{comparison}, where a configuration with $r = 10$ meters, $\phi_{max} = 180$ degrees, and 8 possible orientations is used in the Teknikhuset's map. Our algorithm in this case performs much closer to the offline one (with 4 orientations): a total of 20 sensing positions with an estimated coverage time of 20.68 minutes. While increasing the number of orientations in the offline approach would result in a dramatic increase of the search space, in our approach this can be done without worsening too much the time required to compute the next best smell. Note also that the low exploration time is mostly due to our usage of a dynamic angle for sensing operations. In fact, $\phi_{max}$ is an upper bound in our approach, while in \cite{asif} the scanning angle $\phi$ used for each sensing operation is always set equal to $\phi_{max}$.

Using a machine equipped with an Intel Core i7-Q740 1.73 GHz and 4 GB of memory, simulating the online coverage on the Teknikhuset's map with 4 orientations takes 0.106 seconds, while a test with 8 orientations takes 0.146 seconds, compared with the offline's computing time of about 1 second obtained with 4 orientations.
Tests on the randomly generated grids (Fig.~\ref{graph_grids}) confirm the difference in computing time required by the two approaches, with the offline one requiring from less than a second for small grids to about 10 minutes for 90x90 grids, being able to compute covering paths for 70x70 grids within 5 minutes. With our approach, computing time spans from less than a second for small grids to a maximum of 40 seconds for 90x90 grids, being able to greedily compute covering paths for 70x70 grids in about 20 seconds. In general, it is clear that our online approach can scale much better than the offline one to large environments.

\begin{table}[]
\centering
\resizebox{\linewidth}{!}{%
\begin{tabular}{l|c|c|c|l}
\cline{2-4}
\multicolumn{1}{c|}{} & \cellcolor[HTML]{FFC702}Offline 90/4 & \cellcolor[HTML]{FFC702}Next-Best-Smell 90/4 & \cellcolor[HTML]{FFC702}Next-Best-Smell 180/8 &  \\ \cline{1-4}
\multicolumn{1}{|l|}{\cellcolor[HTML]{FFC702}Sensing operations} & 17 & 40 & 20 &  \\ \cline{1-4}
\multicolumn{1}{|l|}{\cellcolor[HTML]{FFC702}Travel time (s)} & 472 & 519 & 424 &  \\ \cline{1-4}
\multicolumn{1}{|l|}{\cellcolor[HTML]{FFC702}Scanning time (s)} & 629 & 1146 & 816 &  \\ \cline{1-4}
\multicolumn{1}{|l|}{\cellcolor[HTML]{FFC702}Exploration time (m)} & 18.35 & 27.78 & 20.68 &  \\ \cline{1-4}
\end{tabular}}
\caption{Comparison between offline and online approaches. Two configurations are shown for the our online approach: 4 orientations with 90 degrees maximum scan angle and 8 orientations with 180 degrees maximum scan angle.}
\label{comparison}
\end{table}

\section{Related Work}\label{state}

Our work is embedded in the fields of mobile robot olfaction and exploration strategies, addressing one of the topics of the former and exploiting techniques developed in the latter, hence we briefly survey related works from those areas.

\subsection{Robot Olfaction}

In the past two decades mobile robot olfaction has been attracting increasing attention because of the flexibility a mobile platform can offer in a dangerous tasks like gas detection \cite{kowadlo,ishida}.
Early works concerned the combination of mobile robots with \textit{in situ} sensors for mapping gas distribution \cite{bennetts,bennetts2} and for leak detection \cite{soldan,gas2}.

A big contribution was the introduction of Tunable Diode Laser Absorption Spectroscopy (TDLAS) which enables remote measurements up 30 or more meters \cite{bennetts2}. The development of efficient sensing strategies for remote gas sensors is a rather new field and is the topic of this paper.
Atanasov \textit{et al.} in \cite{atanasov} propose a non-greedy algorithm to plan a sensing path under the assumption of linear Gaussian sensing models. The approach works offline and has sub-optimality guarantees. However, the formulation to target tracking provided in \cite{atanasov} is not immediately applicable to our gas sensing problem.
An offline approach towards area coverage in the context of gas sensing is proposed by Arain \textit{et al.} in \cite{asif}, as extension of the work of Tamioka \textit{et al.} \cite{tomioka}, based on the combination of the \textit{Art Gallery Problem} and the \textit{Travelling Salesman Problem}. First, they identify the set of positions from which the robot is able to cover (perceive) the whole environment and, then, they calculate the minimum path connecting all these positions.
In this way Arain \textit{et al.} are able to find a close-to-optimal solution. In previous sections, we extensively compared our approach with that of~\cite{asif}.

\subsection{Exploration Strategies}

Majority of strategies to explore initially unknown environments make decisions greedily during task execution for the different situations the robot encounters and are often called Next-Best-View (NBV). Usually, in NBV systems, candidate locations are chosen in such a way that they are on the frontier between the already explored free space and the unknown one \cite{yamauchi} and they are reachable from the current position of the robot. In \cite{yamauchi}, Yamauchi suggests to use the \textit{travelling cost}, according to which the next best observation location is the nearest one. Gonz\'ales-Ba\~{n}os and Latombe\cite{gonzales} and Stachniss and Burgard \cite{stachniss} combine the travelling cost with \textit{information gain}, that is the expected amount of new information about the environment the robot can acquire performing a sensing operation from the candidate location. In their work, Amigoni and Caglioti \cite{caglioti} introduce a technique based on relative entropy, while Tovar \textit{et al.} \cite{tovar} use several criteria and a multiplicative function to obtain a global utility value.
The mentioned strategies define \textit{ad hoc} aggregation methods that combine the values assumed by the criteria considered. To overcome this issue, Amigoni and Gallo \cite{gallo} proposed an approach based on multi-objective optimization, which is more theoretically founded. This idea has been further developed with the adoption of Multi-Criteria Decision Making in~\cite{mcdm}.

Also some theoretical studies have addressed the problem of online coverage. For example, Gabriel and Rimon \cite{Gabriely2003197} consider the coverage problem in an environment with a grid of squared cells (whose size is equal to the size of the tool covering the area), which is initially unknown. Their online algorithm incrementally builds a spanning tree according to the current knowledge of the environment and generates a covering path whose length is bounded. However, some non fully realistic assumptions (like that on the size of the cells) make these results difficult to apply to our problem.

The contribution of this paper is to use MCDM for originally addressing and solving the gas detection task with an online approach that considers novel criteria with respect to those used in exploration.

\section{Conclusions} \label{conclusion}

In this paper we addressed the problem of planning coverage paths for gas detection with a mobile robot equipped with a TDLAS remote gas sensor.
The solution we presented follows an online approach and uses Multi-Criteria Decision Making (MCDM) to define a Next-Best-Smell coverage greedy strategy that chooses the next sensing pose by combining different criteria. Differently from the exploration settings in which MCDM has been employed so far, sensing time is an important factor in our application and we account for it in a specific criterion, beyond the more usual criteria related to distance traveled and information gain. Experimental evidence shows the potential of the proposed approach in finding effective coverage strategies for mobile robots equipped with gas sensors of limited range and visibility, especially when complete map coverage is not necessary, but coverage of large parts of environments is sufficient. The Next-Best-Smell approach obtains performance similar to that of a state-of-the-art offline approach, but it is much more computationally light and can easily scale to large environments.

The most interesting direction of future research is to extend the Next-Best-Smell approach to unknown environments. We preliminarily ran some tests in simulation without giving the robot any \textit{a priori} information about the environment, and assuming that the range of the laser sensor used to detect obstacles is at least twice the range of the TDLAS sensor. These tests completed successfully with an estimated coverage time at most 10\% longer than in the known environment case.
Other interesting extensions include the use of multiple robots, the use of other criteria for selecting the next best smelling pose (like the quality of gas detection), and, more generally, the investigation of the bounds on the performance of online approaches with respect to the optimal performance of offline approaches.


%



%

\bibliography{biblio}
\bibliographystyle{plain}

%




\end{document}